\documentclass{article} 
\usepackage{iclr2027_conference,times}

\usepackage{amsmath,amsfonts,bm}

\def\eqref#1{equation~\ref{#1}}

\def\1{\bm{1}}

\DeclareMathAlphabet{\mathsfit}{\encodingdefault}{\sfdefault}{m}{sl}
\SetMathAlphabet{\mathsfit}{bold}{\encodingdefault}{\sfdefault}{bx}{n}

\usepackage{url}
\usepackage{amsmath}

\usepackage{enumitem}
\usepackage{graphicx}
\usepackage{float}
\usepackage{wrapfig}
\usepackage{pifont}
\usepackage{multirow}
\usepackage{booktabs}
\usepackage{colortbl}
\usepackage{hyperref}
\usepackage{threeparttable}
\usepackage{bbm}
\usepackage{makecell}
\usepackage{subfigure}

\newcommand{\Checkmark}{\ding{51}}
\newcommand{\XSolidBrush}{\ding{55}}
\newcommand{\Partialmark}{\ensuremath{\triangle}}

\newcommand{\proj}{BioDyad}

\title{{\proj}: Synchronize Biomedical Discovery and Machine Learning Engineering}

\author{Xingbo Du$^{1}$, Fadli Aulawi Al Ghiffari$^{1}$, Leonard Song$^{2}$, Loka Li$^{1}$, Duzhen Zhang$^{1}$,\\
\textbf{Zixiao Wang$^{1}$, Xiuying Chen$^{1}$, Le Song$^{1}$}\\
{\normalfont\small $^{1}$Mohamed bin Zayed University of Artificial Intelligence, Abu Dhabi, UAE}\\
{\normalfont\small $^{2}$The Westminster Schools, Atlanta, GA, USA}
}

\iclrfinalcopy 
\begin{document}

\maketitle
\lhead{Preprint}

\begin{abstract}
Agentic biomedical machine learning (ML) draws on complementary advances in
biomedical evidence acquisition and executable program search. Existing systems
connect aspects of these capabilities, but coordinating them throughout program
search remains challenging. New evidence must guide candidate construction,
execution outcomes must inform subsequent discovery and reuse, and validation
demands must fit the search budget. We introduce \emph{{\proj}}, which couples
biomedical discovery and ML engineering through two hierarchies within
Monte Carlo graph search.
Its scientific hierarchy combines prior biomedical guidance with iterative
discovery, then links biomedical plans to execution outcomes in memory for reuse
across candidates. Its engineering hierarchy moves candidate programs from
smoke execution, through train/validation evaluation, to full-data retraining.
We evaluate {\proj} on the 76-task BioXArena benchmark under a two-hour
per-task budget with three matched LLM backends. It achieves the highest
penalized all-task score and task success rate among four agent methods and a
one-shot baseline under each backend. These results support coordinating biomedical discovery and ML
engineering to integrate external knowledge into executable programs across
heterogeneous biomedical tasks.
\end{abstract}

\section{Introduction}

Machine learning (ML) agents are increasingly capable of constructing and refining
complete learning programs rather than selecting models in
isolation~\citep{chan2025mle,jiang2025aide,du2026mlevolve}. This
capability is especially consequential in biomedicine, where a useful program
must translate task-specific biomedical evidence into executable modeling
decisions across heterogeneous data modalities and task
families~\citep{miller2025bioml,li2026bioxarena}. Knowing biomedical methods
does not itself provide access to new measurements or private institutional
resources. Using such resources requires identifying relevant evidence,
integrating it into candidate programs, and checking its empirical value.
As search proceeds, newly acquired evidence and execution outcomes can change
which representations, models, and validation strategies are
appropriate~\citep{huang2025biomni,jin2025stella}. Autonomous biomedical ML
therefore requires biomedical discovery and machine learning engineering (MLE) to advance together.

Existing work broadly follows two complementary research streams with different
emphases. General-purpose MLE agents develop executable programs through
iterative search, execution feedback, and experience
reuse~\citep{jiang2025aide,du2026mlevolve,zhu2026toward}. Biomedical agents
acquire and interpret domain evidence through specialized resources, tools,
and reasoning workflows~\citep{huang2025biomni,jin2025stella,
roohani2025biodiscoveryagent}. These streams already intersect in systems such
as VCHarness, which combines biomedical profiling and biological models with
program search and feedback memory~\citep{cheng2026harnessing}. Its node
lifecycle centers on full fine-tuning and does not explicitly adapt exploration
to the remaining budget.

Table~\ref{tab:motivation} distinguishes the capabilities involved in
coordinating these two streams. Biomedical discovery acquires task-specific
evidence, and biomedical memory retains discovery plans and execution outcomes
across candidates. Program search compares executable alternatives, while
staged evaluation and budget-adaptive search govern how candidates are tested
and exploration effort is allocated. However, sustaining this coordination
throughout program search remains challenging: new biomedical evidence must
guide candidate revisions, while execution feedback informs subsequent discovery
under a fixed budget. Integrating such evidence can introduce errors and
additional validation demands, making execution cost part of the coordination
problem.

\begin{table}[tb!]
\caption{Capabilities of representative agentic systems for biomedical ML
program discovery.}
\vspace{3pt}
\label{tab:motivation}
\centering
\resizebox{\textwidth}{!}{%
\begin{threeparttable}
\begin{tabular}{llccccc}
\toprule
\multirow{2}{*}[-0.25em]{\textbf{Type}} &
\multirow{2}{*}[-0.25em]{\textbf{Method}} &
\multicolumn{2}{c}{\textbf{Scientific hierarchy}} &
\multicolumn{3}{c}{\textbf{Search and engineering}} \\
\cmidrule(lr){3-4}\cmidrule(lr){5-7}
& & \textbf{Bio. discovery}
& \textbf{Bio. memory}
& \textbf{Program search}
& \textbf{Staged eval.}
& \textbf{Budget adapt.} \\
\midrule
Harness & BioXArena harness~\citep{li2026bioxarena}
& \XSolidBrush & \XSolidBrush & \XSolidBrush & \XSolidBrush & \XSolidBrush \\
\addlinespace[1pt]
\rowcolor{black!4}
\multirow[t]{3}{*}{Bio. agents}
& Biomni~\citep{huang2025biomni}
& \Checkmark & \XSolidBrush & \XSolidBrush & \XSolidBrush & \XSolidBrush \\
\rowcolor{black!4}
& STELLA~\citep{jin2025stella}
& \Checkmark & \XSolidBrush & \XSolidBrush & \XSolidBrush & \XSolidBrush \\
\rowcolor{black!4}
& VCHarness$^*$~\citep{cheng2026harnessing}
& \Partialmark & \Partialmark & \Checkmark & \XSolidBrush & \XSolidBrush \\
\addlinespace[1pt]
\multirow[t]{2}{*}{MLE agents}
& MLEvolve~\citep{du2026mlevolve}
& \XSolidBrush & \XSolidBrush & \Checkmark & \XSolidBrush & \Checkmark \\
& ML-Master 2.0~\citep{zhu2026toward}
& \XSolidBrush & \XSolidBrush & \Checkmark & \XSolidBrush & \Checkmark \\
\midrule
\rowcolor{black!4}
Coupled & \textbf{{\proj} (ours)}
& \Checkmark & \Checkmark & \Checkmark & \Checkmark & \Checkmark \\
\bottomrule
\end{tabular}
\begin{tablenotes}
\item[*] VCHarness profiles biomedical data and curated biological models; its
memory primarily stores program feedback.
\end{tablenotes}
\end{threeparttable}
}
\vspace{-10pt}
\end{table}

To address this gap, we introduce \emph{{\proj}}, a hierarchical agentic search framework that
coordinates biomedical discovery and MLE throughout program search.
Within a shared Monte Carlo graph search (MCGS) backbone~\citep{du2026mlevolve},
its \emph{scientific hierarchy} combines task-level guidance with discovery
tailored to the candidate being expanded. Biomedical memory links these plans
to execution outcomes, allowing later candidates to reuse productive
integrations and revisit failed attempts. Its \emph{engineering hierarchy}
evaluates programs through progressively more demanding execution stages,
providing feasibility and predictive feedback on biomedical proposals during
search.

We evaluate {\proj} on the heterogeneous BioXArena benchmark~\citep{li2026bioxarena} under a two-hour
per-task budget. Under each of three matched LLM backends, {\proj} attains
the highest penalized all-task score and success rate among four baselines.
Component ablations on two domains support the combined use of biomedical
discovery, memory, and staged execution.

This paper makes four main contributions.

1) We introduce \textbf{{\proj}}, a hierarchical agentic search framework
that coordinates \textbf{biomedical discovery and ML engineering} throughout
program search under a fixed budget. Biomedical evidence and execution
outcomes jointly guide subsequent candidate revisions.
    
2) We develop a \textbf{scientific hierarchy} that combines prior guidance,
candidate-level discovery, and plan-outcome memory to support iterative
biomedical resource integration. This enables later candidates to reuse and
revise prior integration decisions.

3) We develop an \textbf{engineering hierarchy} that advances
knowledge-augmented programs from smoke execution through train/validation
evaluation to full-data retraining and submission. These stages separate
feasibility screening, predictive assessment, and final model fitting.

4) We establish {\proj}'s aggregate advantage on the 76-task BioXArena
benchmark through a comparison under three matched LLM backends.
Component ablations examine the contribution of the coupled hierarchies, and
extended-search trajectories characterize progress within {\proj} as more
time becomes available.

\section{Related Work}
\textbf{Agents for Biomedical Discovery.}
Biomedical agents combine LLM reasoning with domain tools, data, and experimental
context. Biomni unifies biomedical resources through a code
interface~\citep{huang2025biomni}, while STELLA adds evolving reasoning templates
and autonomous tool creation~\citep{jin2025stella}. Targeted systems close
narrower loops: BioDiscoveryAgent uses knowledge and prior outcomes to propose
genetic perturbations~\citep{roohani2025biodiscoveryagent}, while the Virtual Lab
coordinates specialized agents with human feedback to design and validate
nanobodies~\citep{swanson2025virtual}. VCHarness directly targets biomedical ML
program development by coupling a coding agent with biological foundation models
for perturbation-response search~\citep{cheng2026harnessing}. Together, these
systems establish knowledge use, tool orchestration, experimental design, and
domain-specific program search. {\proj} centers their coupling by revising
candidate-level findings and retaining them with execution outcomes to guide
later search.

\textbf{Agentic Search for Executable ML Programs.}
Classical AutoML optimizes model components and hyperparameters within
predefined search spaces, as exemplified by auto-sklearn~\citep{feurer2015efficient}.
LLM agents instead search executable code: MLAgentBench and MLGym support
iterative experimentation~\citep{huang2024mlagentbench,nathani2025mlgym},
MLE-Bench evaluates competition-scale performance~\citep{chan2025mle}, and AIDE
performs tree search over code solutions~\citep{jiang2025aide}. For longer
horizons, MLEvolve combines progressive MCGS with retrospective cross-branch
memory~\citep{du2026mlevolve}, while ML-Master 2.0 uses hierarchical cognitive
caching~\citep{zhu2026toward}. BioML-bench and BioXArena extend end-to-end
evaluation to heterogeneous biomedical tasks and
modalities~\citep{miller2025bioml,li2026bioxarena}. These systems establish
program search, memory, and empirical feedback. {\proj} builds on MLEvolve's
search to integrate external biomedical resources through
candidate-level discovery and staged, budget-aware validation.

\begin{figure*}[tb!]
\centering
\includegraphics[width=\textwidth]{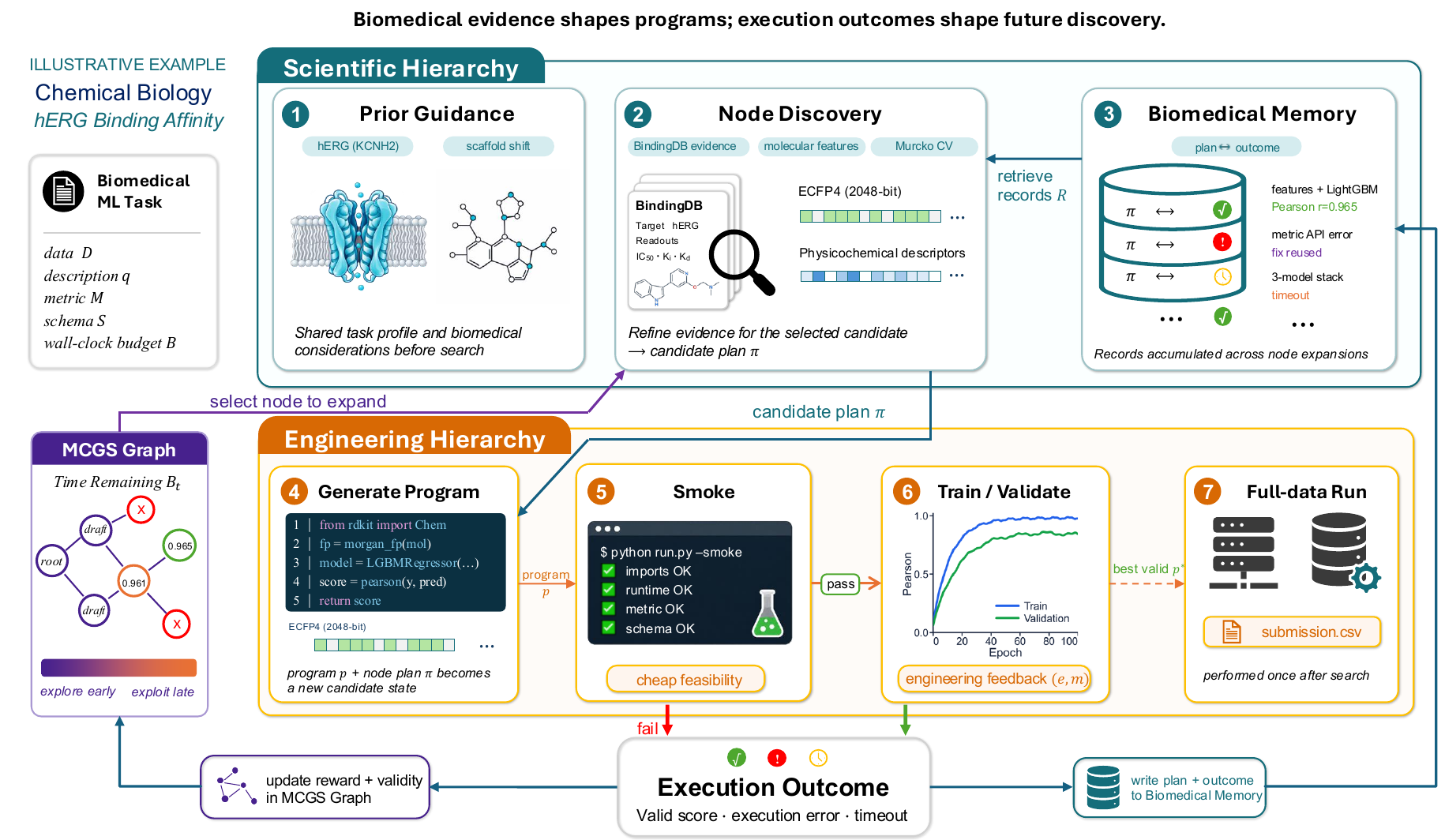}
\vspace{-5pt}
\caption{{\proj} couples biomedical discovery and ML engineering within MCGS.
The scientific hierarchy combines prior guidance, node-level discovery, and
biomedical memory; the engineering hierarchy advances programs through smoke
execution, train/validation evaluation, and full-data runs. Biomedical evidence
guides candidate expansion, while execution feedback updates the search graph
and memory within the wall-clock budget. The \emph{hERG Binding Affinity} task
from BioXArena~\citep{li2026bioxarena} provides an illustrative example.}
\label{fig:overview}
\vspace{-10pt}
\end{figure*}

\section{Problem Formulation}

We consider a biomedical ML task
$\tau=(\mathcal{D},q,M,\mathcal{S},B)$, where $\mathcal{D}$ denotes the public
task data, $q$ the natural-language task description, $M$ the evaluation
metric, $\mathcal{S}$ the required submission schema, and $B$ the total
wall-clock budget. The search space contains complete programs that load data, train and validate a predictive model, and produce a submission
conforming to $\mathcal{S}$. Let $m_\tau(p)$ be the validation score of a valid
program and let $V_\tau(p)\in\{0,1\}$ indicate whether $p$ executes and satisfies
the task contract. If $d_\tau\in\{-1,+1\}$ encodes the optimization direction of
$M$, the objective is
\begin{equation}
p^\star=\arg\max_{p\in\mathcal{P}_B}\ d_\tau m_\tau(p)
\quad\text{subject to}\quad V_\tau(p)=1,
\end{equation}
where $\mathcal{P}_B$ is the set of candidates explored within budget $B$.
The program and its supporting biomedical evidence are coupled: execution
outcomes inform subsequent resource use, while incorporating new evidence can
increase validation cost and fragility. {\proj}
addresses it by evolving biomedical evidence across search and
advancing the resulting programs through staged execution.

\section{{\proj}}
\textbf{Overview.}
Fig.~\ref{fig:overview} summarizes the coupled search loop. At iteration $t$,
{\proj} maintains an MCGS program graph $G_t$ and a biomedical memory
$\mathcal{K}_t$. Each graph node $v$ indexes a candidate state
\begin{equation}
x_v=(p_v,\pi_v,e_v,m_v),
\end{equation}
where $p_v$ is its program, $\pi_v$ its biomedical node plan, $e_v$ its
execution outcome, and $m_v$ its validation score; $e_v$ and $m_v$ remain
pending until execution. Given a parent node $v_t$, the scientific hierarchy
retrieves relevant records and produces a plan for a child node $v'$, after
which the engineering hierarchy instantiates and evaluates its program. Here,
$g_\tau$ denotes the task-level prior guidance:
\begin{equation}
\label{eq:coupled-expansion}
\begin{split}
&R_t=\mathrm{Retrieve}(\mathcal{K}_t,x_{v_t}),\qquad \qquad \qquad
\pi_{v'}=\mathrm{Discover}(\tau,g_\tau,x_{v_t},R_t), \\
&p_{v'}=\mathrm{Generate}(p_{v_t};g_\tau,\pi_{v'},R_t),\qquad
(e_{v'},m_{v'})=\mathrm{Engineer}(p_{v'}).
\end{split}
\end{equation}

The child produces $G_{t+1}$, while its plan and observed outcome produce
$\mathcal{K}_{t+1}$. Biomedical evidence therefore changes the programs, and execution changes the evidence available to later expansions.

\subsection{Scientific Hierarchy}

The scientific hierarchy controls how biomedical evidence is initialized,
revised, and retained. It combines shared prior guidance, node-level discovery,
and a memory that links candidate plans to their computational consequences.

\subsubsection{Biomedical Discovery}\label{sec:biomedical-discovery}

\begin{figure}[t]
\centering
\resizebox{0.88\textwidth}{!}{%
\includegraphics[width=\linewidth]{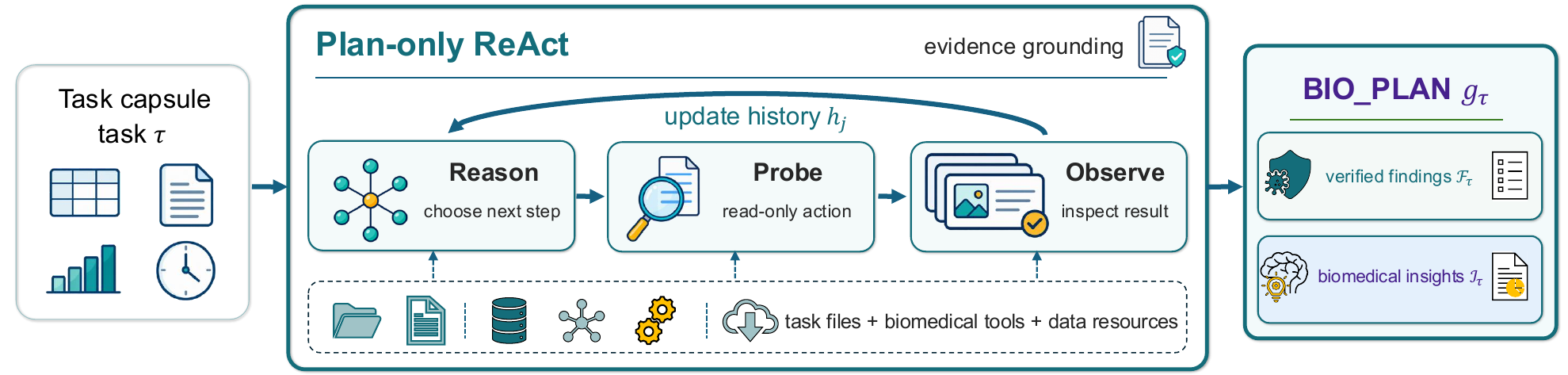}
}
\vspace{-5pt}
\caption{ReAct-based prior biomedical guidance. A plan-only agent alternates
reasoning with read-only probes to compile verified data findings and biomedical insights.}
\label{fig:prior-guidance}
\vspace{-10pt}
\end{figure}

\textbf{Prior guidance.}
Before search, {\proj} runs a plan-only biomedical agent to establish shared
prior guidance $g_\tau$ (Fig.~\ref{fig:prior-guidance}). Its ReAct-style
loop~\citep{yao2023react} alternates reasoning with lightweight, read-only
probes of task files and, when relevant, retrieved biomedical tools and data
resources.

Let $h_{j-1}$ denote the reasoning-observation history before probe step $j$.
The biomedical planner $\mathcal{A}_{\mathrm{bio}}$ selects action $a_j$,
executes it through the available tool interface $\mathcal{T}$, and appends the
observation $o_j$:
\begin{equation}
a_j\sim\mathcal{A}_{\mathrm{bio}}(\cdot\mid\tau,h_{j-1}),\qquad
o_j=\mathcal{T}(a_j),\qquad
h_j=h_{j-1}\oplus(a_j,o_j).
\end{equation}
When the agent terminates the loop, it compiles the accumulated evidence into
\begin{equation}
g_\tau=\operatorname{Compile}(h_J)
=\bigl(\mathcal{F}_\tau,\mathcal{I}_\tau\bigr),
\end{equation}
where $\mathcal{F}_\tau$ contains verified data findings and
$\mathcal{I}_\tau$ contains biomedical implications supported by those
findings. The resulting \texttt{BIO\_PLAN} conditions subsequent node discovery
without prescribing a model family or acting as an MCGS operator.

\textbf{Discovery during node expansion.}
Before generating each draft, improvement, or debugging node, {\proj} invokes
a biomedical discovery agent with the task, prior guidance, selected parent,
and retrieved memory. The agent may inspect relevant data and resources, then
returns a candidate-specific \texttt{NODE\_PLAN} $\pi_{v'}$ that binds the
biomedical rationale to an implementable modeling and validation strategy.
Since discovery occurs after earlier candidates have been evaluated, it can
refine supported directions, abandon contradicted ones, or branch to a distinct
hypothesis. Biomedical knowledge is therefore revised throughout program
evolution rather than injected only at initialization.

\subsubsection{Biomedical Memory}\label{sec:memory}

Biomedical memory extends MLEvolve's experience reuse~\citep{du2026mlevolve}
to plans informed by biomedical discovery and their execution outcomes.
For each MCGS node, biomedical memory stores a compact record
\begin{equation}
k_v=(\pi_v,s_v,e_v,m_v,\Delta m_v,\ell_v),\qquad \mathcal{K}_{t+1}=\operatorname{Upsert}(\mathcal{K}_t,k_{v'}).
\end{equation}

Here, $s_v$ is the search stage, $\Delta m_v$ is the score change relative to
the parent, and $\ell_v$ marks the record as successful or failed.
The plan is inserted before code generation and the same keyed record is
updated after execution.
Both successful and failed attempts remain available for retrieval, alongside
a contrasting viable direction when useful. Subsequent discovery can reuse
or revise earlier attempts to integrate external biomedical knowledge using
their recorded execution outcomes.
The record connects what a candidate attempted with what happened during
execution and how its score changed. This preserves the distinction between
an integration error and an executable change with little predictive benefit,
giving subsequent discovery a basis for different revisions.

\subsection{Engineering Hierarchy}
\label{sec:engineering-hierarchy}

The engineering hierarchy turns knowledge-augmented programs into comparable
feedback without committing the full search budget to every candidate. It
separates feasibility, comparative evaluation, and final retraining: \textit{1) Smoke Execution.}
Each generated program first undergoes static checks and smoke execution. A
reduced-cost smoke run retains the candidate's feature representation, model,
and output format to test whether the proposed integration is executable.
The candidate advances only if it completes without an exception, reports the
required validation score, and produces a submission artifact. Otherwise, its
failure is recorded and control returns to MCGS rather than entering a nested
repair loop.
\textit{2) Train/Validation Evaluation.}
Smoke-passing candidates run their intended train/validation program. The
evaluator extracts the task metric, verifies the required artifact and schema,
and returns validity and a score $m_v$. These outcomes update both MCGS and the
plan-outcome record in biomedical memory.
This stage assesses predictive quality under the candidate's intended training
and validation settings, supplying the scores used to compare executable
alternatives.
\textit{3) Full-Data Retraining and Submission.}
After search, {\proj} selects the best valid program using its pre-retraining
validation score. A final variant retrains the same approach on all labeled
public data and writes the submission.

\subsection{Coupling the Hierarchies}

At each search iteration, MCGS~\citep{du2026mlevolve} selects a node to expand
according to its execution history and the remaining wall-clock budget. The
selected parent and retrieved plan-outcome records condition the next
biomedical plan, which guides construction of the child program in
Eq.~\ref{eq:coupled-expansion}. Staged execution then updates two states with
different roles. Validity and validation performance inform subsequent node
selection in the search graph, while biomedical memory retains the rationale
associated with each outcome. Later discovery can therefore revise resource
integration using feedback from earlier candidates.

Implementation settings are detailed in Appdx.~\ref{app:implementation-settings},
and the MCGS backbone and search configuration in Appdx.~\ref{app:mcgs}.

\section{Experiments}

\subsection{Benchmark and Evaluation Protocol}

\textbf{Benchmark.}
BioXArena~\citep{li2026bioxarena} contains 76 held-out prediction tasks across nine biomedical domains:
sequence, single-cell, structure, network biology, chemical biology,
perturbation dynamics, phenotype-disease, imaging, and text-integrated biology.
The benchmark is deliberately heterogeneous, with 46 of the 76 tasks combining multiple modality families. Each
public task capsule contains a task description, public train/test inputs,
native-format assets when required, a sample submission, and a task-specific
metric; test labels and evaluators remain hidden.

\textbf{Execution protocol.}
We follow the official BioXArena interface. An agent receives the public task
capsule in a no-internet sandbox, trains predictive models under a two-hour
wall-clock limit and the prescribed CPU/GPU allocation, and writes
\texttt{submission.csv}. BioXArena permits additional biomedical knowledge
through a locally mounted data cache~\citep{li2026bioxarena};
{\proj} uses local Biomni data and tools~\citep{huang2025biomni} during
biomedical discovery
(Appdx.~\ref{app:biomni}). All phases, including prior planning, candidate
discovery and evaluation, and final retraining and submission, are included
in the two-hour limit. The primary comparison fixes the task interface,
hardware allocation, and time limit across five methods under each LLM backend. Extended-budget and repeated-run protocols are specified
separately below.

\begin{figure}[tb!]
\centering
\includegraphics[width=\textwidth]{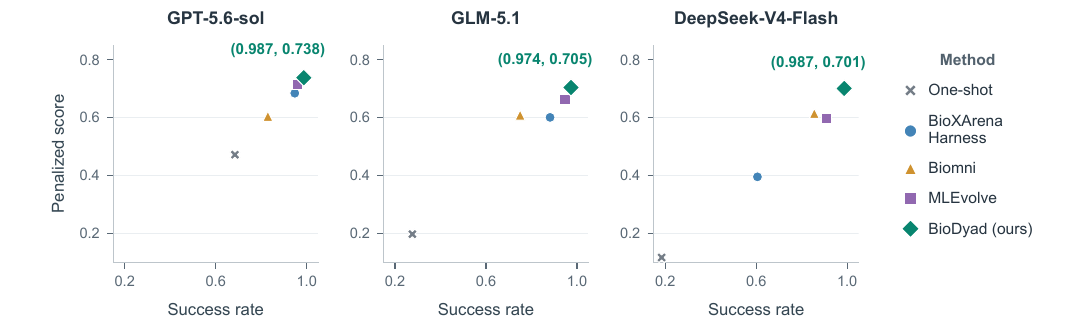}
\vspace{-10pt}
\caption{Success rate (x-axis) and penalized all-task score (y-axis) under three LLM backends.
Each panel compares four agent methods with a two-hour per-task budget and a
one-shot baseline retaining only the first attempt, evaluated on the same 76 tasks.}
\label{fig:main-results-a}
\vspace{-10pt}
\end{figure}

\textbf{Methods and backends.}
The primary comparison evaluates {\proj}, MLEvolve~\citep{du2026mlevolve},
Biomni~\citep{huang2025biomni}, and BioXArena Harness~\citep{li2026bioxarena} with each of
GPT-5.6-sol~\citep{openai2026gpt56sol}, GLM-5.1~\citep{zai2026glm51}, and
DeepSeek-V4-Flash Preview~\citep{xu2026deepseek}.
In this comparison, Biomni accesses its own data lake and tools, whereas
MLEvolve and BioXArena Harness do not. The resource-matched control, MLEvolve w/ discovery, is
reported in the component ablations (Table~\ref{tab:ablations}).
Fig.~\ref{fig:main-results-a} and Table~\ref{tab:backend-control} additionally
include a one-shot baseline for each backend, retaining only the first attempt
per task from the corresponding BioXArena Harness run and scoring failures as zero.
For broader context,
Appdx.~\ref{sec:reported-systems} compares {\proj} with the results
presented on BioXArena's leaderboard, retaining each baseline's reported
backend configuration.

\textbf{Scoring.}
BioXArena maps each task metric to $[0,1]$, with larger values indicating better
performance. Let $s_{a,i}$ be the score of agent $a$ on task $i$,
with $s_{a,i}=0$ for a failed or missing submission. 
We report the penalized mean score $\bar{s}_a$ over all
76 tasks and the success rate $R_a$, defined as the fraction of tasks
with successful execution and valid submissions.

\begin{table}[tb!]
\centering
\caption{Domain breakdowns of four agent methods and a one-shot baseline under three matched LLM backends on the 76
BioXArena tasks. Scores are penalized
average scores; higher is better. Avg. is weighted by task
count across all 76 tasks. \textbf{Bold} and \underline{underlining} indicate the
highest and second-highest distinct displayed scores, respectively, within each backend
and column.}
\resizebox{\textwidth}{!}{%
\begin{threeparttable}
\label{tab:backend-control}
\footnotesize
\begin{tabular}{llcccccccccc}
\toprule
\textbf{LLM backend} & \textbf{Method} & \textbf{Seq} & \textbf{SC} &
\textbf{Str} & \textbf{Net} & \textbf{Chem} & \textbf{Pert} &
\textbf{Phen} & \textbf{Img} & \textbf{Text} & \textbf{Avg.} \\
\midrule
\multirow{5}{*}{GPT-5.6-sol} & One-shot & 0.416 & 0.288 & 0.727 & 0.458 & 0.576 & 0.291 & 0.508 & 0.500 & 0.546 & 0.472 \\
 & BioXArena Harness & 0.663 & 0.786 & \textbf{0.834} & 0.458 & \textbf{0.928} & 0.517 & \textbf{0.606} & \textbf{0.714} & 0.637 & 0.685 \\
 & Biomni & 0.773 & 0.652 & 0.591 & 0.490 & 0.690 & 0.404 & 0.431 & \underline{0.702} & 0.656 & 0.605 \\
 & MLEvolve & \textbf{0.790} & \underline{0.798} & 0.794 & \textbf{0.527} & 0.922 & \underline{0.597} & 0.575 & 0.673 & \textbf{0.720} & \underline{0.715} \\
 & \textbf{{\proj} (ours)} & \underline{0.776} & \textbf{0.861} & \underline{0.815} & \underline{0.521} & \underline{0.926} & \textbf{0.776} & \underline{0.594} & 0.666 & \underline{0.671} & \textbf{0.738} \\
\midrule
\multirow{5}{*}{GLM-5.1} & One-shot & 0.187 & 0.102 & 0.236 & 0.062 & 0.556 & 0.121 & 0.374 & 0.139 & 0.030 & 0.198 \\
 & BioXArena Harness & 0.649 & 0.753 & \textbf{0.824} & \textbf{0.511} & 0.894 & 0.336 & 0.487 & 0.513 & 0.393 & 0.601 \\
 & Biomni & 0.580 & 0.669 & \underline{0.796} & 0.487 & 0.912 & 0.618 & 0.310 & \underline{0.639} & 0.467 & 0.609 \\
 & MLEvolve & \underline{0.713} & \textbf{0.791} & 0.729 & \textbf{0.511} & \underline{0.919} & \underline{0.621} & \underline{0.584} & 0.439 & \textbf{0.610} & \underline{0.662} \\
 & \textbf{{\proj} (ours)} & \textbf{0.776} & \underline{0.771} & 0.729 & \underline{0.509} & \textbf{0.932} & \textbf{0.737} & \textbf{0.600} & \textbf{0.700} & \underline{0.552} & \textbf{0.705} \\
\midrule
\multirow{5}{*}{DeepSeek-V4} & One-shot & 0.085 & 0.095 & 0.095 & 0.137 & 0.229 & 0.111 & 0.072 & 0.000 & 0.244 & 0.117 \\
 & BioXArena Harness & 0.623 & 0.423 & 0.095 & \underline{0.508} & 0.506 & 0.183 & 0.398 & 0.353 & 0.408 & 0.396 \\
 & Biomni & 0.673 & 0.618 & 0.720 & 0.431 & 0.870 & 0.571 & \underline{0.510} & \underline{0.621} & \underline{0.504} & \underline{0.615} \\
 & MLEvolve & \textbf{0.739} & \underline{0.761} & \underline{0.733} & 0.442 & \underline{0.902} & \underline{0.578} & 0.440 & 0.386 & 0.306 & 0.596 \\
 & \textbf{{\proj} (ours)} & \underline{0.717} & \textbf{0.791} & \textbf{0.789} & \textbf{0.510} & \textbf{0.909} & \textbf{0.717} & \textbf{0.600} & \textbf{0.646} & \textbf{0.608} & \textbf{0.701} \\
\bottomrule
\end{tabular}
\begin{tablenotes}[flushleft]
\footnotesize
\item[] \textit{Domains:} Seq, sequence; SC, single-cell; Str, structure;
Net, network biology; Chem, chemical biology; Pert, perturbation dynamics;
Phen, phenotype-disease; Img, imaging; Text, text-integrated.
\end{tablenotes}
\end{threeparttable}
}
\vspace{-10pt}
\end{table}

\subsection{Main Results}
\label{sec:main-results}

{\proj} achieves the highest penalized all-task score and execution success
rate under each of the three LLM backends (Fig.~\ref{fig:main-results-a}).
This aggregate advantage extends across
most domains: {\proj} attains the highest or second-highest distinct reported
score in eight, eight, and nine of nine domains with GPT-5.6-sol, GLM-5.1, and DeepSeek-V4, respectively (Table~\ref{tab:backend-control}).
Perturbation dynamics is a consistent strength: {\proj} leads this domain
under every backend. The leading methods in other domains vary with the
backend, showing that the aggregate result combines a consistent domain
advantage with broad competitiveness across task families. {\proj} also
successfully completes nearly all tasks under each backend.
Additionally, all four agent methods exceed the corresponding one-shot
average under every backend. The one-shot success rates of 0.684, 0.276,
and 0.184, respectively, highlight execution reliability as a limitation
alongside predictive quality.
Biomni, despite using the same data lake, stays within 0.605-0.615.

More specifically, {\proj} improves the penalized score over the strongest matched alternative
by 0.023, 0.042, and 0.086
with GPT-5.6-sol, GLM-5.1, and DeepSeek-V4, respectively
(computed before rounding). It also narrows the penalized-score gap between
GPT-5.6-sol and DeepSeek-V4 from 0.119 with MLEvolve to 0.037.
With DeepSeek-V4, {\proj} reaches 0.701, approaching MLEvolve with
GPT-5.6-sol (0.715). These gains support coupling biomedical discovery
with execution throughout search across the tested backends. The component
ablations below examine the contributions of the two hierarchies.

\begin{wraptable}{r}{0.45\textwidth}
\vspace{-20pt}
\centering
\caption{Repeated-run robustness on the fixed 18-task subset with GPT-5.6-sol.
Values are mean $\pm$ sample standard deviation (SD) across three independent runs.}
\label{tab:repeat-summary}
\vspace{3pt}
\resizebox{0.45\textwidth}{!}{%
\begin{tabular}{lcc}
\toprule
\textbf{Method} & \textbf{Penalized score} & \textbf{Success rate} \\
\midrule
BioXArena Harn. & $0.699\,{\scriptstyle\pm0.005}$ & $\mathbf{0.944}\,{\scriptstyle\pm0.000}$ \\
Biomni & $0.647\,{\scriptstyle\pm0.012}$ & $0.889\,{\scriptstyle\pm0.000}$ \\
MLEvolve & $0.688\,{\scriptstyle\pm0.019}$ & $0.907\,{\scriptstyle\pm0.032}$ \\
{\proj} & $\mathbf{0.711}\,{\scriptstyle\pm0.002}$ & $\mathbf{0.944}\,{\scriptstyle\pm0.000}$ \\
\bottomrule
\end{tabular}
}
\vspace{-10pt}
\end{wraptable}
\textbf{Repeated-Run Robustness.}
To assess run-to-run variability across biomedical domains, we fix an
18-task subset comprising the first two tasks in each domain in BioXArena's
Table A4~\citep{li2026bioxarena}. We use GPT-5.6-sol for BioXArena Harness, Biomni, MLEvolve, and {\proj} with three independent runs per method-task
pair under the two-hour budget. Each run starts from a fresh workspace. The task list and
per-run results are in Appdx.~\ref{app:repeated-runs}.
Across three runs with GPT-5.6-sol, {\proj} combines the highest
mean penalized score with a small sample SD across runs
(Table~\ref{tab:repeat-summary}).

\subsection{Component Ablations}
\vspace{-5pt}
\begin{table}[h]
\caption{Component ablations with the GPT-5.4 backend. Scores are penalized
task averages.
Avg. is the task-weighted mean
across all 16 tasks in the two domains; higher is better.}
\vspace{3pt}
\label{tab:ablations}
\centering
\resizebox{\textwidth}{!}{%
\begin{tabular}{llcccccc}
\toprule
\multirow{2}{*}{\textbf{Hierarchy}} &
\multirow{2}{*}{\textbf{Mechanism}} &
\multirow{2}{*}{\textbf{MLEvolve}} &
\textbf{MLEvolve} & \textbf{{\proj}} & \textbf{{\proj}} & \textbf{{\proj}} & \textbf{{\proj}}\\
& & & \textbf{w/ discovery} & \textbf{w/o memory} &\textbf{w/o Sci.} & \textbf{w/o Eng.} & \textbf{(full)} \\
\midrule
\multirow{2}{*}{Sci. hierarchy}
& Bio. discovery & \XSolidBrush & \Checkmark & \Checkmark & \XSolidBrush & \Checkmark & \Checkmark \\
& Bio. memory & \XSolidBrush & \XSolidBrush & \XSolidBrush &\XSolidBrush & \Checkmark & \Checkmark \\
\midrule
Eng. hierarchy
& Staged evaluation & \XSolidBrush & \XSolidBrush & \Checkmark & \Checkmark & \XSolidBrush & \Checkmark \\
\midrule
\multirow{3}{*}[-0.25em]{Performance $\uparrow$}
& Chemical biology & 0.898 & 0.795 & 0.913 & 0.861 & 0.896 & \textbf{0.919} \\
& Phenotype-disease & 0.538 & 0.599 & 0.560 & 0.468 & 0.570 & \textbf{0.606} \\
\cmidrule{2-8}
& Average & 0.718 & 0.697 & 0.736 & 0.665 & 0.733 & \textbf{0.763} \\
\bottomrule
\end{tabular}
}
\vspace{-5pt}
\end{table}

Following BioXArena's scaling analysis~\citep{li2026bioxarena}, we ablate the
scientific and engineering hierarchies on chemical-biology and phenotype-disease
tasks (Table~\ref{tab:ablations}).
The ablations remove either the scientific hierarchy, comprising biomedical
discovery (Sec.~\ref{sec:biomedical-discovery}) and memory (Sec.~\ref{sec:memory}),
or staged evaluation (Sec.~\ref{sec:engineering-hierarchy}). We also remove
biomedical memory alone, retaining discovery and staged evaluation.
MLEvolve w/ discovery adds discovery without biomedical memory or staged
evaluation, matching the full framework's biomedical resources and discovery
capabilities.
To decouple component analysis from the three main evaluation backends,
all ablation variants use GPT-5.4~\citep{openai2026gpt54} under the same
two-hour task budget.

The full framework achieves the highest average score (0.763), compared with
0.718 for MLEvolve and 0.697 for MLEvolve w/ discovery
(Table~\ref{tab:ablations}). Biomedical resources thus pay off only when
coupled with memory or staged evaluation. Removing biomedical memory or staged evaluation
reduces the average to 0.736 and 0.733, respectively. These comparisons support
contributions from both memory and staged evaluation when biomedical discovery
is retained.

Discovery alone improves phenotype-disease prediction but reduces
chemical-biology performance relative to MLEvolve. 
Staged evaluation without the scientific hierarchy yields 0.665, below MLEvolve.
Adding discovery alone therefore does not reproduce the benefit of the
complete framework. The two domains favor different partial configurations,
yet both reach their highest score when discovery, memory, and staged
evaluation are used together.
Candidate-level diagnostics in
Appdx.~\ref{app:engineering-ablation} and \ref{app:scientific-ablation}
characterize the corresponding screening and execution outcomes.

\subsection{Case Studies}
\label{sec:case-studies}

\begin{figure}[tb!]
\centering
\includegraphics[width=\linewidth]{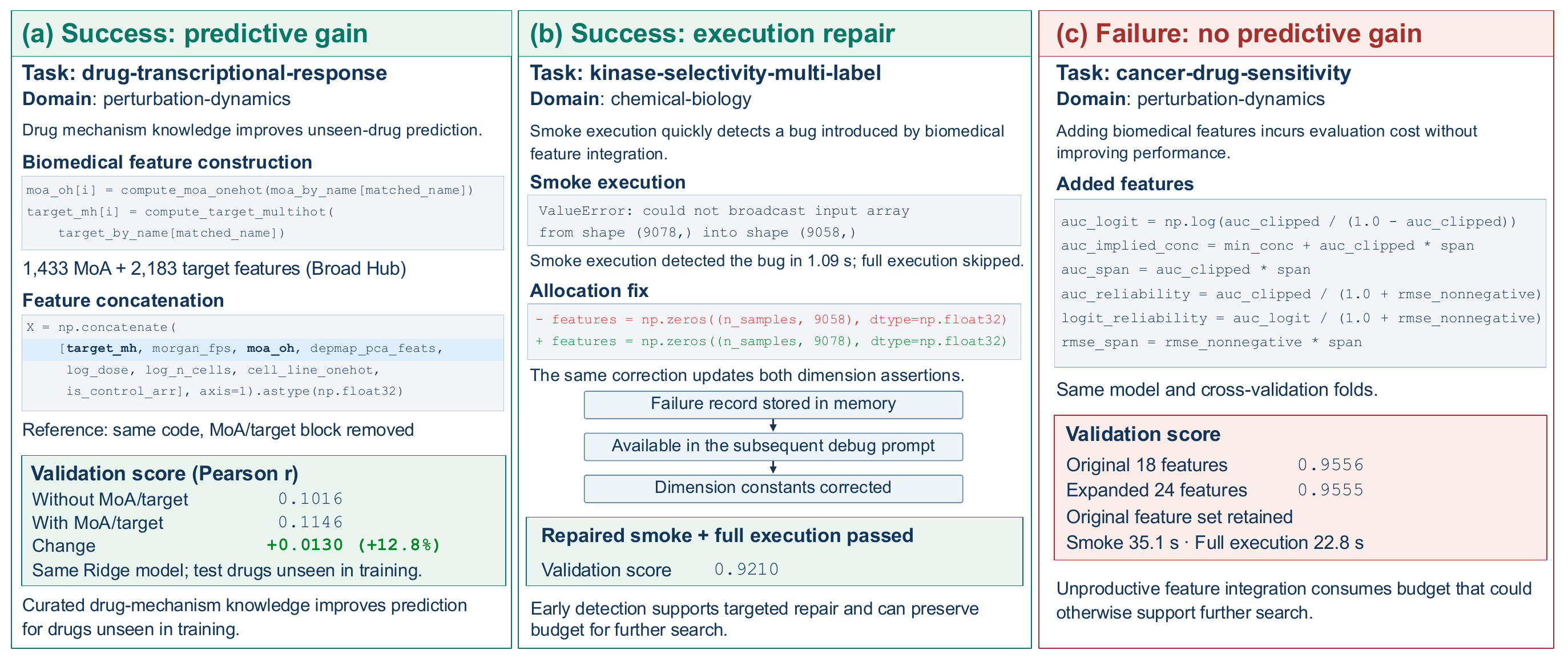}
\vspace{-15pt}
\caption{Biomedical feature integration in {\proj}: (a) predictive gain,
(b) smoke-guided execution repair, and (c) no predictive gain despite successful
execution. Scores are task-specific validation metrics; full execution denotes
candidate train/validation evaluation.}
\label{fig:case_studies}
\vspace{-10pt}
\end{figure}

Fig.~\ref{fig:case_studies} illustrates why biomedical discovery and staged
execution are complementary. Biomedical evidence can improve predictive
representations, but its integration can also introduce execution errors.
Smoke execution provides early feedback for repair, while train/validation
evaluation tests whether executable changes improve prediction.
Appdx.~\ref{app:case-studies} details the tasks and candidate changes.

\textbf{Successful case: Predictive gain.}
On the \emph{drug-transcriptional-response} task~\citep{srivatsan2020massively},
test drugs are unseen during training and identified only by name.
Mechanism-of-action (MoA) and target annotations from the Drug Repurposing
Hub~\citep{corsello2017drug} raised validation Pearson $r$ from 0.1016 to
0.1146 with the same model.

\textbf{Successful case: Smoke-guided repair.}
On the \emph{kinase-selectivity-multi-label} task~\citep{mendez2019chembl}, additional molecular features introduced
a dimension mismatch. Smoke execution detected the error in 1.09~s and skipped
full execution. The recorded feedback guided a subsequent repair that retained
the added features and restored successful execution.

\textbf{Failed case: No predictive gain.}
On the \emph{cancer-drug-sensitivity} task~\citep{iorio2016landscape}, added biomedical features executed successfully
but did not improve prediction, so the original representation was retained.
The unproductive extension consumed search budget.

\subsection{Budget Extension within BioDyad}
\label{sec:time-budget-analysis}

We examine whether extending search helps {\proj} discover better candidate
programs with GPT-5.6-sol. We use the tasks as in the component ablations. At each two-hour checkpoint of a 12-hour
run, we retrospectively record the highest normalized hidden-test score among
candidates completed by that time. Reading all checkpoints from one trajectory
per task tracks how the candidate pool improves as search continues.

\begin{table}[tb!]
\centering
\caption{Retrospective best-so-far normalized test scores during 12-hour
{\proj} runs with GPT-5.6-sol. Each checkpoint averages task-level maxima
over all eight tasks in a domain, using the same runs as
Appdx.~\ref{app:budget-score-analysis}. Gain columns report absolute
differences computed before rounding.}
\vspace{5pt}
\label{tab:budget-score}
\resizebox{0.92\textwidth}{!}{%
\begin{tabular}{l|cccccc|ccc}
\toprule
\multirow{2}{*}[-0.25em]{\textbf{Domain}} &
\multicolumn{6}{c|}{\textbf{Best-so-far test score}} &
\multicolumn{3}{c}{\textbf{Score gain}} \\
\cmidrule(lr){2-7}\cmidrule(lr){8-10}
& \textbf{2 h} & \textbf{4 h} & \textbf{6 h} & \textbf{8 h} &
\textbf{10 h} & \textbf{12 h} & \textbf{2-4 h} & \textbf{4-12 h} &
\textbf{2-12 h} \\
\midrule
Chemical biology & 0.927 & 0.932 & 0.936 & 0.936 & 0.936 & 0.937 & +0.006 & +0.004 & +0.010 \\
Phenotype-disease & 0.640 & 0.663 & 0.688 & 0.690 & 0.690 & 0.690 & +0.022 & +0.027 & +0.050 \\
\bottomrule
\end{tabular}
}
\vspace{-5pt}
\end{table}

Mean best observed normalized test scores rise by 0.010 in chemical biology
and 0.050 in phenotype-disease from 2 to 12 hours
(Table~\ref{tab:budget-score}). Both domains gain during \mbox{2-4 h} and \mbox{4-12 h},
with slightly larger later gains in phenotype-disease. Extended search thus
uncovers better candidates after the standard budget.

\begin{figure}[tb!]
\centering
\includegraphics[width=\linewidth]{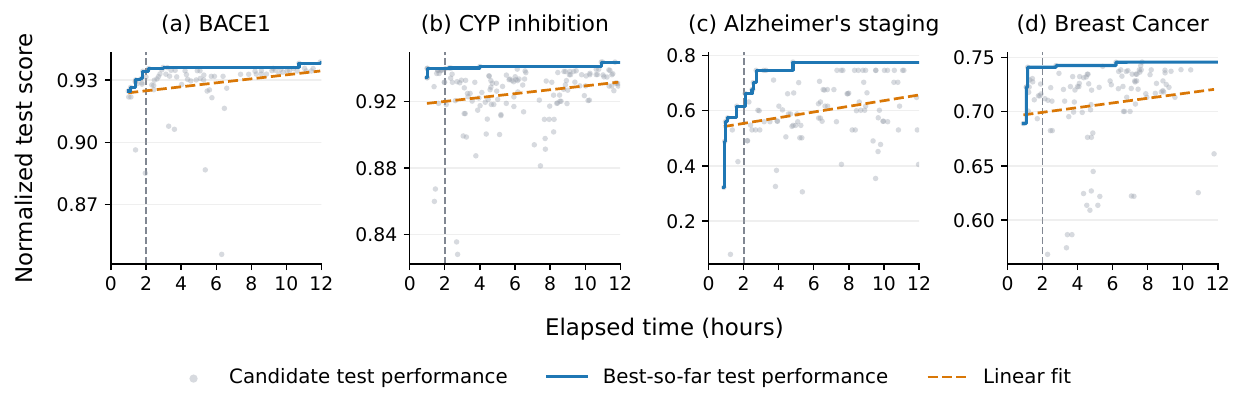}
\vspace{-15pt}
\caption{Candidate test scores during 12-hour searches:
(a) BACE1 binding affinity, (b) CYP inhibition, (c) Alzheimer's disease staging, (d) breast-cancer subtype. Gray points show candidate scores; blue lines show
retrospective maxima; orange dashed lines show linear fits against time.}
\vspace{-10pt}
\label{fig:visualization}
\end{figure}

Fig.~\ref{fig:visualization} shows how these gains emerge from candidate
exploration. Alzheimer's disease staging improves after the two-hour checkpoint
and plateaus at about five hours, showing that better candidates can emerge
beyond the initial search period. BACE1 and CYP inhibition reach high scores early
and improve more modestly thereafter. Breast-cancer subtype shows a similar
early plateau despite substantial variation among later candidates. The
positive slopes of ordinary least-squares linear fits in all four examples indicate an upward
trend in average candidate test performance over time, despite variation
among individual candidates.

Returns to additional search are thus task-dependent. The domain means
aggregate gains accumulated at different stages of these trajectories.
Appdx.~\ref{app:budget-score-analysis} provides the complete
task-level test-score trajectories and checkpoints.

\section{Conclusion}

We introduced {\proj}, a hierarchical agentic search framework that
coordinates biomedical discovery and ML engineering under a fixed budget.
Its scientific hierarchy combines prior guidance, candidate-level discovery,
and biomedical memory, while its engineering hierarchy supplies feasibility
and predictive feedback through staged evaluation.
On 76 BioXArena tasks, {\proj} achieves the highest penalized all-task score
and success rate under each of three matched LLM backends.
Ablations in chemical biology and phenotype-disease support the joint use of
discovery, memory, and staged evaluation, beyond adding discovery alone.
Together, these results support using execution outcomes to guide how
biomedical evidence is integrated and reused throughout program search.
The current study has three limitations: (i) biomedical discovery depends
on the quality, coverage, and accessibility of external resources;
(ii) prior guidance and candidate-level discovery require additional LLM
calls within the available budget; and (iii) evaluation covers supervised
BioXArena tasks, with component and extended-budget analyses restricted to
two domains. Extension to wet-lab or other long-horizon scientific workflows
remains to be evaluated.

\newpage

\subsection*{AI use statement}

We used generative AI tools to polish the language and improve the readability
of the manuscript, and to generate initial drafts of figures. The authors take
full responsibility for the accuracy and integrity of the final manuscript,
including all AI-assisted text and figures.



\bibliography{iclr2027_conference}
\bibliographystyle{iclr2027_conference}

\newpage
\appendix

\section{Comparison with the BioXArena Leaderboard}
\label{sec:reported-systems}

Table~\ref{tab:reported-systems} places {\proj} alongside the agent and
BioXArena harness results presented on BioXArena's leaderboard~\citep{li2026bioxarena}, retaining each
baseline's reported backend configuration. All three {\proj} configurations
exceed the highest reported overall score of 0.666 among the listed baselines.
At least one {\proj} configuration achieves the highest score in eight of
the nine domains. Because backend configurations differ across these entries,
the matched-backend comparison in Section~\ref{sec:main-results} isolates the
framework's contribution.

\begin{table}[tb!]
\caption{Comparison with the penalized score presented on BioXArena's leaderboard across all 76 tasks. \textbf{Avg.} is task-weighted across all
tasks. Baseline scores are reported by BioXArena~\citep{li2026bioxarena}; {\proj} scores use the same
evaluation protocol. Bold marks the highest score in each column. Backends differ across systems (Appdx.~\ref{app:baseline-backends}). }
\vspace{3pt}
\label{tab:reported-systems}
\centering
\resizebox{\textwidth}{!}{%
\begin{tabular}{lcccccccccc}
\toprule
\textbf{Agent} & \textbf{Seq} & \textbf{SC} & \textbf{Str} &
\textbf{Net} & \textbf{Chem} & \textbf{Pert} & \textbf{Phen} &
\textbf{Img} & \textbf{Text} & \textbf{Avg.} \\
\midrule
{\proj} (GPT-5.6-sol) & \textbf{0.776} & \textbf{0.861} & 0.815 & \textbf{0.521} & 0.926 & \textbf{0.776} & 0.594 & 0.666 & \textbf{0.671} & \textbf{0.738} \\
{\proj} (GLM-5.1) & \textbf{0.776} & 0.771 & 0.729 & 0.509 & \textbf{0.932} & 0.737 & \textbf{0.600} & \textbf{0.700} & 0.552 & 0.705 \\
{\proj} (DeepSeek-V4) & 0.717 & 0.791 & 0.789 & 0.510 & 0.909 & 0.717 & \textbf{0.600} & 0.646 & 0.608 & 0.701 \\
\midrule
MLEvolve~\citep{du2026mlevolve} & 0.765 & 0.798 & 0.750 & 0.510 & 0.869 & 0.601 & 0.502 & 0.606 & 0.534 & 0.666 \\
STELLA~\citep{jin2025stella} & 0.677 & 0.559 & 0.481 & 0.497 & 0.878 & 0.555 & 0.574 & 0.651 & 0.641 & 0.613 \\
Biomni~\citep{huang2025biomni} & 0.526 & 0.646 & 0.569 & 0.497 & 0.723 & 0.628 & 0.547 & 0.659 & 0.415 & 0.579 \\
ML-Master 2.0~\citep{zhu2026toward} & 0.582 & 0.773 & 0.337 & 0.509 & 0.904 & 0.401 & 0.411 & 0.601 & 0.342 & 0.547 \\
\midrule
GPT-5.4~\citep{openai2026gpt54} & 0.748 & 0.673 & \textbf{0.836} & 0.507 & 0.866 & 0.391 & 0.501 & 0.601 & 0.565 & 0.636 \\
GLM-5.1~\citep{zai2026glm51} & 0.649 & 0.753 & 0.824 & 0.511 & 0.894 & 0.336 & 0.487 & 0.513 & 0.393 & 0.601 \\
Claude Opus 4.6~\citep{anthropic2026claudeopus46} & 0.665 & 0.496 & 0.726 & 0.509 & 0.605 & 0.416 & 0.507 & 0.656 & 0.565 & 0.572 \\
Gemini 3.1 Pro~\citep{geminiteam2026gemini31pro} & 0.596 & 0.681 & 0.738 & 0.510 & 0.827 & 0.400 & 0.490 & 0.635 & 0.607 & 0.611 \\
Gemma 4 31B~\citep{team2026gemma} & 0.568 & 0.683 & 0.730 & 0.430 & 0.746 & 0.214 & 0.411 & 0.490 & 0.293 & 0.513 \\
Qwen3.6-Plus~\citep{alibabacloud2026qwen36plus} & 0.606 & 0.764 & 0.666 & 0.445 & 0.844 & 0.363 & 0.528 & 0.622 & 0.436 & 0.591 \\
DeepSeek-V3.2~\citep{liu2025deepseek} & 0.256 & 0.521 & 0.399 & 0.434 & 0.365 & 0.144 & 0.287 & 0.507 & 0.262 & 0.355 \\
\bottomrule
\end{tabular}
}
\end{table}

\section{Implementation Details}
\label{app:implementation}

\subsection{Biomedical Resources from Biomni}
\label{app:biomni}

Biomni~\citep{huang2025biomni} provides a biomedical environment comprising
domain-specific tools, curated data resources, and scientific
software.\footnote{\url{https://biomni.stanford.edu/}}
Its tools support operations such as sequence analysis, molecular analysis,
and cell-type annotation; its data resources include expression profiles,
functional annotations, and disease associations; and its software environment
provides executable analysis routines. These resources span heterogeneous
biomedical workflows, so their relevance depends on the task and the candidate
being developed.

In {\proj}, Biomni resources support prior guidance and candidate-level
discovery. The discovery agent uses resource descriptions and data schemas to
identify relevant evidence, inspect compatibility with task inputs, and propose
concrete representation or analysis changes. For example, drug
mechanism-of-action annotations can supply transferable information after drug
names are matched to resource entries. The configured workspace exposes data-lake schemas and
tool/software catalogs for this purpose. Resource use is selective: when no
suitable resource can be matched to the task, the candidate can retain its
existing representation. Proposed changes subsequently pass through staged
execution and validation, with their outcomes recorded for later search.

Staged validation then determines which integrations improve prediction, as
illustrated by the contrasting cases in Appendix~\ref{app:case-studies}.

\subsection{Discovery, Memory, and Execution Settings}
\label{app:implementation-settings}

The settings below complement the scientific and engineering hierarchies in
the main text. Resource catalogs and schemas are described in
Appendix~\ref{app:biomni}.

\begin{center}
\small
\renewcommand{\arraystretch}{1.15}
\begin{tabular}{@{}p{0.18\textwidth}p{0.78\textwidth}@{}}
\toprule
\textbf{Component} & \textbf{Implementation setting} \\
\midrule
Prior guidance & The Biomni A1-based planner runs once before search, with at
most six read-only probe blocks and no candidate training. \\
\addlinespace[4pt]
Memory retrieval & Keyword and embedding rankings are combined through
reciprocal-rank fusion. Each retrieval supplies up to two related successful
or pending plans, two related failures, and one dissimilar successful plan;
duplicates are removed. \\
\addlinespace[4pt]
Smoke execution & Static checks precede a reduced-cost smoke run that preserves
the full program's feature representation, model, and output format. Passing
requires successful execution, a validation score, and a submission artifact. \\
\addlinespace[4pt]
Finalization & The selected program retains its pre-retraining score for model
selection. Its full-data variant is smoke-gated and executed once; if this
fails, the original valid program and submission are retained. \\
\bottomrule
\end{tabular}
\end{center}

\subsection{MCGS Backbone and Search Configuration}
\label{app:mcgs}

\textbf{Search representation and loop.}
Monte Carlo graph search (MCGS) organizes candidate programs as nodes and
iteratively selects, expands, and evaluates them. Following
MLEvolve~\citep{du2026mlevolve}, primary edges record parent-child generation
and carry reward updates back through the selected node's ancestors.
Cross-branch references provide additional context while preserving this
primary ancestry. Selecting the root produces a new draft; selecting an invalid
candidate triggers debugging; selecting a valid candidate triggers improvement.
In {\proj}, the scientific hierarchy supplies the biomedical plan and retrieved
memory for each expansion, and the engineering hierarchy evaluates the resulting
program. Validity and validation improvement determine its reward, which updates
the search graph; its plan and outcome also update biomedical memory.

\textbf{Balancing exploration and refinement.}
The inherited selection strategy combines upper confidence bounds applied to
trees (UCT) with elite-guided sampling. UCT balances observed reward against
exploration of less-visited nodes; elite-guided sampling prioritizes
high-scoring valid candidates. For a visited node $v$, UCT uses
\begin{equation}
u_t(v)=\frac{Q_v}{N_v}+C_t
\sqrt{\frac{\log(N_{\operatorname{pa}(v)}+1)}{N_v}},
\end{equation}
where $Q_v$ is cumulative reward, $N_v$ is visit count,
$\operatorname{pa}(v)$ is the primary parent, and $C_t$ controls exploration.
Unvisited nodes receive priority. Elite-guided selection samples the top-$K$
valid candidates with probability proportional to inverse score rank.

\textbf{Budget-dependent configuration.}
Selection increasingly favors elite candidates.
Let $\rho_t$ denote the elapsed wall-clock budget fraction.
UCT is chosen with probability
\begin{equation}
w_{\mathrm{exp}}(\rho_t)=
\begin{cases}
1, & \rho_t<\rho_s,\\
1-(1-w_{\min})\dfrac{\rho_t-\rho_s}{\rho_e-\rho_s},
& \rho_s\leq\rho_t<\rho_e,\\
w_{\min}, & \rho_t\geq\rho_e,
\end{cases}
\end{equation}
with $\rho_s=0.5$, $\rho_e=0.7$, and $w_{\min}=0.2$; otherwise, selection uses
the elite strategy. The elite pool contains up to five candidates before
$\rho_e$ and three afterward, with at most three and two candidates from each
branch, respectively. These search rules follow the MLEvolve backbone;
{\proj} connects them to biomedical discovery, memory, and staged execution.

\section{Ablation Studies and Search Diagnostics}
\label{app:ablation-diagnostics}
\raggedbottom

Table~\ref{tab:ablations} reports aggregate predictive performance. This
appendix examines the same component ablations at the candidate level, asking
how each component changes the reliability of the programs that search
actually executes. The runs use GPT-5.4, the 16 chemical-biology and
phenotype-disease tasks, and a two-hour task budget. Counts are pooled across
tasks within each configuration. 

\subsection{Diagnostic Setup and Counting Rules}
\label{app:diagnostic-counting}

\emph{Full-execution runs} are candidates that start train/validation
evaluation. Without staged evaluation, every generated candidate enters this
stage directly; with staged evaluation, only candidates that pass static checks
and smoke execution do. Each full-execution run receives one of two terminal
labels. \emph{Valid} runs complete with a finite validation score, no buggy
flag, and the required submission artifact. \emph{Invalid}
runs end with an execution defect, an unparseable score or output record, or
no terminal record, typically due to timeout or interrupted bookkeeping.
Percentages use each configuration's number of full-execution runs as the
denominator. These candidate-level outcomes differ from the task-level success
rate in the main results.

\subsection{Candidate Execution Outcomes}
\label{app:engineering-ablation}

Table~\ref{tab:engineering-full-outcomes} reports execution outcomes for all
six configurations. Because the configurations differ by one component at a
time, the table contains matched pairs for each mechanism: two pairs that add
biomedical discovery, two that add biomedical memory, and three that add
staged evaluation.

\begin{table}[H]
\centering
\caption{Candidate outcomes in the component ablations. Each entry gives a
count and its percentage of that configuration's full-execution runs.}
\label{tab:engineering-full-outcomes}
\resizebox{\textwidth}{!}{%
\begin{tabular}{llcccccc}
\toprule
\multirow{2}{*}{\textbf{Hierarchy}} &
\multirow{2}{*}{\textbf{Mechanism}} &
\multirow{2}{*}{\textbf{MLEvolve}} &
\textbf{MLEvolve} & \textbf{{\proj}} & \textbf{{\proj}} & \textbf{{\proj}} & \textbf{{\proj}}\\
& & & \textbf{w/ discovery} & \textbf{w/o memory} &\textbf{w/o Sci.} & \textbf{w/o Eng.} & \textbf{(full)} \\
\midrule
\multirow{2}{*}{Sci. hierarchy}
& Bio. discovery & \XSolidBrush & \Checkmark & \Checkmark & \XSolidBrush & \Checkmark & \Checkmark \\
& Bio. memory & \XSolidBrush & \XSolidBrush & \XSolidBrush &\XSolidBrush & \Checkmark & \Checkmark \\
\midrule
Eng. hierarchy
& Staged evaluation & \XSolidBrush & \XSolidBrush & \Checkmark & \Checkmark & \XSolidBrush & \Checkmark \\
\midrule
\multirow{3}{*}{Outcomes}
& Full runs & 552 & 672 & 639 & 446 & 668 & 566 \\
& Valid & 304 (55.1\%) & 314 (46.7\%) & 425 (66.5\%) & 319 (71.5\%) & 411 (61.5\%) & 458 (80.9\%) \\
& Invalid & 248 (44.9\%) & 358 (53.3\%) & 214 (33.5\%) & 127 (28.5\%) & 257 (38.5\%) & 108 (19.1\%) \\
\bottomrule
\end{tabular}
}
\end{table}

The complete framework has both the highest valid share (80.9\%) and the
largest number of valid runs (458) among the six configurations, together with
the fewest invalid runs (108).

\textbf{Staged evaluation.}
Adding staged evaluation increases the valid share in all three matched pairs:
from 55.1\% to 71.5\% without the scientific hierarchy (MLEvolve $\rightarrow$
{\proj} w/o Sci.), from 46.7\% to 66.5\% with discovery alone (MLEvolve w/
discovery $\rightarrow$ {\proj} w/o memory), and from 61.5\% to 80.9\% with
discovery and memory ({\proj} w/o Eng. $\rightarrow$ {\proj} full). Absolute
counts show where this gain comes from. In each pair, staged evaluation reduces the
number of full executions (552 $\rightarrow$ 446, 672 $\rightarrow$ 639, and
668 $\rightarrow$ 566) and reduces invalid runs by 40--58\% (248 $\rightarrow$
127, 358 $\rightarrow$ 214, and 257 $\rightarrow$ 108), while the number of
valid runs increases (304 $\rightarrow$ 319, 314 $\rightarrow$ 425, and
411 $\rightarrow$ 458). The engineering hierarchy therefore shifts full-execution
budget away from infeasible candidates without reducing the number of valid
programs available for comparison.

\subsection{Scientific Hierarchy and Execution Reliability}
\label{app:scientific-ablation}

\textbf{Biomedical discovery.}
Adding discovery without memory lowers the valid share in both matched pairs:
from 55.1\% to 46.7\% without staged evaluation (MLEvolve $\rightarrow$
MLEvolve w/ discovery), and from 71.5\% to 66.5\% with it ({\proj} w/o Sci.
$\rightarrow$ {\proj} w/o memory). Without staged evaluation, invalid runs
increase from 248 to 358. This is consistent with the coordination problem
stated in the introduction: integrating biomedical evidence into programs
introduces execution errors and additional validation demands, and discovery
alone does not resolve them.

\textbf{Biomedical memory.}
Adding memory to discovery raises the valid share by a similar margin in both
matched pairs: from 46.7\% to 61.5\% without staged evaluation (MLEvolve w/
discovery $\rightarrow$ {\proj} w/o Eng.), and from 66.5\% to 80.9\% with it
({\proj} w/o memory $\rightarrow$ {\proj} full). In both pairs, memory brings
the valid share above that of the corresponding configuration without the
scientific hierarchy (55.1\% and 71.5\%, respectively). Memory links each
biomedical plan to its execution outcome, so later discovery can revise
integrations that failed rather than propose them again. Memory thus turns
discovery from a net cost in execution reliability into a net gain.

\textbf{Relation to predictive performance.}
Execution reliability alone does not determine task-level performance.
{\proj} w/o Sci. has the second-highest valid share (71.5\%) but the lowest
average score in Table~\ref{tab:ablations} (0.665), whereas {\proj} w/o memory
and w/o Eng. reach higher scores (0.736 and 0.733) with lower valid shares.
Staged evaluation makes candidate programs executable, but predictive gains
additionally require the biomedical content supplied by the scientific
hierarchy. Only the complete framework is best on both measures, combining the
highest valid share with the highest average score (0.763). These diagnostics
complement the task-level ablations: biomedical discovery supplies predictive
content, memory and staged evaluation absorb the execution cost of integrating
it, and the benefit appears when discovery and ML engineering advance together.

\section{Budget-Dependent Score Analysis}
\label{app:budget-score-analysis}

Section~\ref{sec:time-budget-analysis} summarizes budget effects by domain
and illustrates four tasks in Fig.~\ref{fig:visualization}. Here we provide
normalized test-score trajectories and checkpoints for all 16 chemical-biology
and phenotype-disease tasks using GPT-5.6-sol, with one 12-hour run per task.
All analyses use the same fixed set of runs, documented in the accompanying
source-data manifest.

\subsection{Trajectory Reconstruction}
\label{app:budget-trajectories}

For each task, we retain candidates with a finite normalized test score and
a recorded completion time within the 12-hour budget. Elapsed time is measured
from the run-start timestamp to the logged candidate completion. The cumulative
best at a checkpoint is the highest normalized test score among candidates
completed by that time. A score is carried forward until another completed
candidate improves it; the last observed best is retained through 12 hours.
Pearson correlations are mapped from $[-1,1]$ to $[0,1]$ using $(r+1)/2$;
metrics already in $[0,1]$ retain their recorded values.

\begin{figure}[H]
\centering
\includegraphics[width=\linewidth]{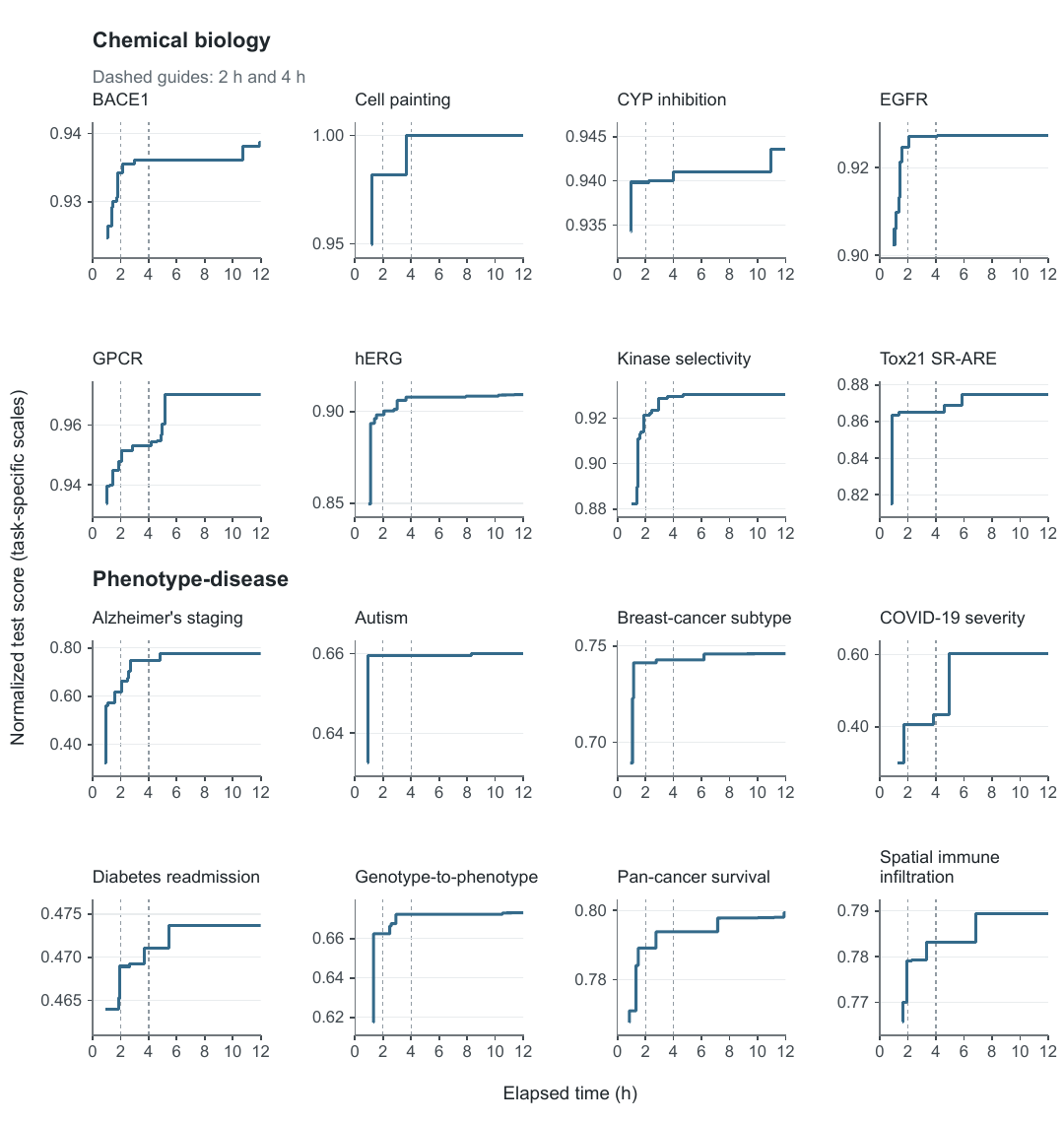}
\caption{Cumulative best normalized test scores for all 16 tasks, using the
same runs as Table~\ref{tab:budget-task-summary}. The upper two rows
show chemical-biology tasks and the lower two show phenotype-disease tasks.
Each task has its own vertical scale. Steps occur at logged candidate
completion; dashed guides mark 2 and 4 hours. No score is shown before the
first completed candidate, and the last observed best is carried forward
to 12 hours.}
\label{fig:budget-score-trajectories}
\end{figure}

Figure~\ref{fig:budget-score-trajectories} shows all 16 trajectories.
These retrospective test-score maxima characterize the evaluated candidate
pool, and all checkpoints for a task belong to the same run.

\subsection{Task-Level Gains and Their Timing}
\label{app:budget-tables}

Table~\ref{tab:budget-task-summary} separates gains between 2 and 4 hours
from gains over the remaining eight hours. All checkpoint values and gains
are computed from the same candidate records as the curves, before
rounding. This task-level view exposes timing differences behind the domain
summaries in Table~\ref{tab:budget-score}.

\begin{table}[H]
\centering
\caption{Normalized test-score checkpoints and gains in the
12-hour traces. CB denotes chemical biology and PD denotes phenotype-disease.
Each checkpoint is a retrospective maximum over completed candidates.
Gains are absolute differences computed before rounding.}
\label{tab:budget-task-summary}
\resizebox{\textwidth}{!}{%
\begin{tabular}{llrrrrr}
\toprule
\textbf{Domain} & \textbf{Task} & \textbf{Test 2 h} &
\textbf{Test 4 h} & \textbf{Test 12 h} &
\textbf{$\Delta$ 2-4 h} & \textbf{$\Delta$ 4-12 h} \\
\midrule
CB & BACE1 Binding Affinity & 0.934 & 0.936 & 0.939 & +0.002 & +0.003 \\
CB & Cell Painting Perturbation & 0.982 & 1.000 & 1.000 & +0.018 & +0.000 \\
CB & CYP Inhibition (Multi-label) & 0.940 & 0.941 & 0.944 & +0.001 & +0.003 \\
CB & EGFR Binding Affinity & 0.925 & 0.927 & 0.927 & +0.003 & +0.000 \\
CB & GPCR Binding (Multi-class) & 0.948 & 0.953 & 0.970 & +0.005 & +0.017 \\
CB & hERG Binding Affinity & 0.898 & 0.908 & 0.909 & +0.010 & +0.001 \\
CB & Kinase Selectivity (Multi-label) & 0.921 & 0.930 & 0.930 & +0.008 & +0.001 \\
CB & Tox21 SR-ARE & 0.865 & 0.865 & 0.875 & +0.000 & +0.010 \\
\midrule
PD & Alzheimer's Disease Staging & 0.617 & 0.746 & 0.776 & +0.130 & +0.030 \\
PD & Autism Diagnosis & 0.660 & 0.660 & 0.660 & +0.000 & +0.000 \\
PD & Breast Cancer Subtype & 0.741 & 0.743 & 0.746 & +0.002 & +0.003 \\
PD & COVID-19 Severity Classification & 0.407 & 0.433 & 0.603 & +0.026 & +0.170 \\
PD & Diabetes Readmission & 0.469 & 0.471 & 0.474 & +0.002 & +0.003 \\
PD & Genotype-to-Phenotype & 0.662 & 0.672 & 0.673 & +0.010 & +0.001 \\
PD & Pan-Cancer Survival Prediction & 0.789 & 0.794 & 0.799 & +0.005 & +0.005 \\
PD & Spatial Immune Infiltration & 0.779 & 0.783 & 0.789 & +0.004 & +0.006 \\
\bottomrule
\end{tabular}
}
\end{table}

The timing of improvement differs across tasks. Cell painting and Alzheimer's
staging gain mainly between 2 and 4 hours, whereas COVID-19 severity and GPCR
gain mainly after 4 hours. The genotype-to-phenotype run also gains
mainly between 2 and 4 hours, with a smaller subsequent increase. Autism's
gain after 2 hours is below 0.001. The full task set therefore exposes timing
and magnitude differences that a domain average or a small set of examples
can obscure.

\clearpage
\flushbottom

\section{Additional Evaluation Protocols}

\subsection{Repeated-Run Task Selection and Aggregation}
\label{app:repeated-runs}

The repeated-run study fixes the first two
tasks in each of the nine domains in Table A4 of BioXArena~\citep{li2026bioxarena}.
Table~\ref{tab:repeat-tasks} lists the exact 18 task identifiers. This
deterministic, domain-balanced rule fixes the subset independently of method
performance.
All four methods use the same list, GPT-5.6-sol, task inputs, evaluator,
hardware allocation, and two-hour limit in each of three independent runs.
Mutable search state, generated artifacts, and method-specific memory are
reset between task-runs; configured static resources remain fixed.

\begin{table}[tb!]
\centering
\caption{Fixed repeated-run subset: the first two tasks per domain in
BioXArena Table A4, preserving its within-domain order.}
\label{tab:repeat-tasks}
\footnotesize
\vspace{3pt}
\begin{tabular}{lll}
\toprule
\textbf{Domain} & \textbf{First task} & \textbf{Second task} \\
\midrule
Sequence & \texttt{gene-\allowbreak tissue-\allowbreak expression} & \texttt{isoform-\allowbreak expression} \\
Single-cell & \texttt{batch-\allowbreak integration} & \texttt{cell-\allowbreak type-\allowbreak from-\allowbreak expression} \\
Structure & \texttt{complex-\allowbreak structure-\allowbreak evaluation} & \texttt{enzyme-\allowbreak commission-\allowbreak prediction} \\
Network biology & \texttt{gene-\allowbreak disease-\allowbreak association} & \texttt{go-\allowbreak function-\allowbreak multi-\allowbreak label} \\
Chemical biology & \texttt{bace1-\allowbreak binding-\allowbreak affinity} & \texttt{cell-\allowbreak painting-\allowbreak perturbation} \\
Perturbation dynamics & \texttt{cancer-\allowbreak drug-\allowbreak sensitivity} & \texttt{crispr-\allowbreak perturbation-\allowbreak prediction} \\
Phenotype-disease & \texttt{alzheimers-\allowbreak disease-\allowbreak staging} & \texttt{autism-\allowbreak diagnosis} \\
Imaging & \texttt{amos-\allowbreak organ-\allowbreak segmentation} & \texttt{drug-\allowbreak moa-\allowbreak prediction} \\
Text-integrated & \texttt{biomedical-\allowbreak figure-\allowbreak vqa} & \texttt{dna-\allowbreak enzyme-\allowbreak function} \\
\bottomrule
\end{tabular}
\end{table}

For method $a$ and repetition $r\in\{1,2,3\}$, we aggregate all 18 tasks:
\begin{equation}
\bar{s}^{(r)}_a=\frac{1}{18}\sum_{i=1}^{18}s^{(r)}_{a,i},\qquad
R^{(r)}_a=\frac{1}{18}\sum_{i=1}^{18}c^{(r)}_{a,i}.
\end{equation}
Here, $c^{(r)}_{a,i}=1$ if the task program executes successfully and produces
a valid submission, and $0$ otherwise. All 18 tasks remain in the denominator.
Table~\ref{tab:repeat-summary} reports the mean and sample SD of each metric
across the three independent runs. These summaries are calculated before
rounding the run-level values reported in Table~\ref{tab:three-run}.

\begin{table}[tb!]
\centering
\caption{Per-run results on the fixed 18-task subset with GPT-5.6-sol.
Score is the penalized mean over all 18 tasks; success rate is the successful
task count divided by 18. Summary entries report mean $\pm$ sample standard
deviation (SD) across three independent runs.}
\label{tab:three-run}
\vspace{3pt}
\begin{tabular}{llcccc}
\toprule
\textbf{Method} & \textbf{Metric} & \textbf{Run 1} & \textbf{Run 2} &
\textbf{Run 3} & \textbf{Mean $\pm$ SD} \\
\midrule
\multirow{2}{*}{BioXArena Harness} & Score & 0.695 & 0.705 & 0.699 & $0.699{\scriptstyle\pm0.005}$ \\
& Success rate & 0.944 & 0.944 & 0.944 & $0.944\,{\scriptstyle\pm0.000}$ \\
\multirow{2}{*}{Biomni} & Score & 0.637 & 0.660 & 0.644 & $0.647{\scriptstyle\pm0.012}$ \\
& Success rate & 0.889 & 0.889 & 0.889 & $0.889{\scriptstyle\pm0.000}$ \\
\multirow{2}{*}{MLEvolve} & Score & 0.709 & 0.683 & 0.672 & $0.688{\scriptstyle\pm0.019}$ \\
& Success rate & 0.944 & 0.889 & 0.889 & $0.907{\scriptstyle\pm0.032}$ \\
\multirow{2}{*}{{\proj}} & Score & 0.712 & 0.708 & 0.712 & $0.711{\scriptstyle\pm0.002}$ \\
& Success rate & 0.944 & 0.944 & 0.944 & $0.944{\scriptstyle\pm0.000}$ \\
\bottomrule
\end{tabular}
\end{table}

\subsection{Backend Configurations of Reported Systems}
\label{app:baseline-backends}

Table~\ref{tab:baseline-backends} records the framework-backend assignments
for the reported-system reference in Section~\ref{sec:reported-systems}.
These are the original BioXArena configurations, distinct from the comparison in Sec.~\ref{sec:main-results}. For direct LLM
baselines, the method name identifies the backend.

\begin{table}[tb!]
\caption{Original backends of BioXArena's reported agent systems, retained for
the overall reference experiment.}
\vspace{3pt}
\label{tab:baseline-backends}
\centering
\begin{tabular}{lp{0.62\textwidth}}
\toprule
\textbf{Method} & \textbf{LLM backend} \\
\midrule
Biomni & Claude Sonnet 4~\citep{anthropic2025claudesonnet4} \\
STELLA & Dev/Tool-Creation: Claude Sonnet 4.6~\citep{anthropic2026claudesonnet46}; \\
& Manager/Critic: Gemini 3.1 Pro~\citep{geminiteam2026gemini31pro} \\
MLEvolve & Gemini 3.1 Pro~\citep{geminiteam2026gemini31pro} \\
ML-Master 2.0 & DeepSeek-V4-Pro~\citep{xu2026deepseek} \\
\bottomrule
\end{tabular}
\end{table}

VCHarness~\citep{cheng2026harnessing} is outside the two-hour quantitative
comparison because its fine-tuning-centered node lifecycle targets multi-day
execution; compressing that lifecycle would change the method being compared.



\subsection{Case Study Details}
\label{app:case-studies}

The cases in Section~\ref{sec:case-studies} were reconstructed from candidate
programs, execution outputs, biomedical plans, and memory records. Scores are
internal validation metrics, distinct from the normalized hidden-test scores
in the benchmark comparison. The first two cases illustrate predictive gain
and successful execution repair; the third illustrates feature integration
that consumes budget without improving prediction.

\subsubsection{Case (a): Successful Case-Predictive Gain}

\textbf{Task and rationale.}
The \emph{drug-transcriptional-response} task~\citep{srivatsan2020massively}
predicts the expression change of 5,000 genes for each drug, cell line, and
dose condition. Each drug is identified only by name, together with its dose,
cell line, cell count, and a control indicator. The 104 training drugs and 84
test drugs do not overlap, so drug identity cannot transfer from training to
test. With GLM-5.1, the prior guidance of {\proj} noted that generalization to
unseen drugs therefore requires drug properties. It found MoA and target
annotations in the Drug Repurposing Hub~\citep{corsello2017drug} for 70 of the
104 training drugs and 58 of the 84 test drugs, with 40 MoA classes shared
across the two sets. The rationale was that drugs sharing a mechanism may
induce similar transcriptional programs.

\textbf{What changed.}
The program matched drug names to Hub entries and encoded 1,433 MoA one-hot
features and 2,183 target multi-hot features. It concatenated these blocks
with Morgan fingerprints, cell-line features derived from DepMap, log dose,
log cell count, a cell-line one-hot encoding, and the control indicator, and
fitted a Ridge regression model. Prior guidance introduced the annotations
from the first draft onward. The reference removes only the MoA/target block
from the same program, retaining all other features, the Ridge model, and the
drug-grouped validation, in which validation drugs are unseen during fitting.

\textbf{Outcome and interpretation.}
Mean per-gene Pearson correlation increased from 0.1016 to 0.1146 (+0.0130,
or 12.8\% relative). The Hub provides pharmacological annotations rather than
transcriptional measurements, so the gain reflects transfer of responses from
training drugs with shared mechanisms, information that drug names alone do
not provide.

\subsubsection{Case (b): Successful Case-Smoke-Guided Repair}

\textbf{Task and rationale.}
The \emph{kinase-selectivity-multi-label} task~\citep{mendez2019chembl} predicts a compound's activity
across multiple kinase targets from its molecular representation. The candidate
added Morgan count fingerprints and an Avalon fingerprint to existing molecular
features. Count fingerprints encode substructure multiplicity, while Avalon
provides a complementary structural encoding. The intended benefit was a richer
representation of compounds for the per-kinase classifiers.

\textbf{What changed and failed.}
With GLM-5.1, the generated program concatenated these feature blocks but
miscalculated their total dimension: it allocated 9,058 columns for a
9,078-dimensional vector. Smoke execution encountered the mismatch during
feature fusion and stopped before full execution. The error was recorded in
biomedical memory and appeared with its traceback in the subsequent debug
prompt. The repaired candidate changed only the allocation and two dimension
assertions, retaining the feature computations, ExtraTrees classifiers, and
hyperparameters.

\textbf{Outcome and interpretation.}
The faulty candidate was rejected after 1.09~s of smoke execution, and full
execution was skipped. After repair, the smoke stage took 150.8~s of wall-clock
time, followed by 616.4~s of full execution. The candidate produced a submission
and achieved macro ROC-AUC of 0.9210 under three-fold stratified validation.
Early feedback thus supported a focused repair that preserved the proposed
molecular representation.

\subsubsection{Case (c): Failed Case-No Predictive Gain}

\textbf{Task and rationale.}
The \emph{cancer-drug-sensitivity} task~\citep{iorio2016landscape} predicts log half-maximal inhibitory
concentration for drug-cell-line pairs. The reference program already used
biomedical measurements summarizing dose-response curves, concentration bounds,
and curve-fitting error, together with drug and cancer-related descriptors.
{\proj} proposed six additional nonlinear features to make relationships among
these measurements explicit, including transformations of curve area and
combinations with concentration range and fitting error.

\textbf{What changed.}
Using GPT-5.6-sol, the candidate expanded the input from 18 to 24 features.
The program evaluated both representations within the same full
execution, using the same histogram-based gradient-boosting regressor,
hyperparameters, random seed, and three-fold splits grouped by cell line.
Selection used Spearman correlation over pooled out-of-fold predictions.

\textbf{Outcome and interpretation.}
The expanded representation scored 0.9555, compared with 0.9556 for the
original, which was retained. The tree model likely already captured these
nonlinear relationships from the original measurements. Excluding discovery and code generation, the trial
required 35.1~s in the smoke stage and 22.8~s in full execution.
The failed feature extension therefore consumed budget that could
otherwise support further search, illustrating that adding biomedical features
does not necessarily improve performance.

\end{document}